%% file: AUDI_paper_copy.tex
\documentclass[10pt, conference]{IEEEtran}

\def\ps@headings{%
\def\@oddhead{\mbox{}\scriptsize\rightmark \hfil \thepage}%
\def\@evenhead{\scriptsize\thepage \hfil \leftmark\mbox{}}%
\def\@oddfoot{}%
\def\@evenfoot{}}
\makeatother
\usepackage[pdftex,dvips]{color,graphicx}
\usepackage{comment}
\usepackage[table,xcdraw]{xcolor}
\usepackage{amssymb}
\usepackage{amsmath}
\usepackage{amsthm}
\usepackage{enumerate}
\usepackage{graphicx}
\usepackage{fancyhdr}
\usepackage{epstopdf}
\usepackage{lscape}
\usepackage[hidelinks]{hyperref}
\usepackage{setspace}
\usepackage{afterpage}
\usepackage[ruled,vlined]{algorithm2e}
\usepackage{subcaption}
\usepackage{array, booktabs}
\usepackage{epigraph}
\usepackage{footnote}
\makesavenoteenv{tabular}
\makesavenoteenv{table}
\makesavenoteenv{minipage}
\usepackage{float}
\usepackage{multirow}
\usepackage[font=small,labelfont=bf]{caption}
\usepackage{bbm}
\usepackage[pass]{geometry}
\usepackage[italian,english]{babel}
\usepackage{enumitem}
\input{newcommands.tex}
\newcommand{\etal}{et al. \!}

\IEEEoverridecommandlockouts

\begin{document}
\title{Driving Behavior Analysis through CAN Bus Data in an Uncontrolled Environment}

\author{\IEEEauthorblockN{
Umberto Fugiglando\IEEEauthorrefmark{1},
Emanuele Massaro\IEEEauthorrefmark{1},
Paolo Santi\IEEEauthorrefmark{1}\IEEEauthorrefmark{3},
Sebastiano Milardo\IEEEauthorrefmark{1}\IEEEauthorrefmark{2},
Kacem Abida\IEEEauthorrefmark{5}, \\
Rainer Stahlmann \IEEEauthorrefmark{4},
Florian Netter \IEEEauthorrefmark{4}, and
Carlo Ratti\IEEEauthorrefmark{1}
}

\IEEEauthorblockA{
	\IEEEauthorrefmark{1}MIT Senseable City Lab, Cambridge, MA, USA}
\IEEEauthorblockA{
	\IEEEauthorrefmark{3}Istituto di Informatica e Telematica del CNR, Pisa, Italy}
\IEEEauthorblockA{
	\IEEEauthorrefmark{2}University of Palermo, Palermo, Italy}
\IEEEauthorblockA{
	\IEEEauthorrefmark{4}AUDI AG}
\IEEEauthorblockA{
	\IEEEauthorrefmark{5}VW Group Electronics Research Laboratory \\
		email: \{umbertof, emassaro, psanti, milardo, ratti\}@mit.edu,\\ rainer.stahlmann@audi.de, kacem.abida@vw.com}
		}

\maketitle

\begin{abstract}

Cars can nowadays record several thousands of signals through the CAN bus technology and potentially provide real-time information on the car, the driver and the surrounding environment. This paper proposes a  new method for the analysis and classification of driver behavior using a selected subset of CAN bus signals, specifically gas pedal position, brake pedal pressure, steering wheel angle, steering wheel momentum, velocity, RPM, frontal and lateral acceleration. Data has been collected in a completely uncontrolled experiment, where 64 people drove 10 cars for or a total of over 2000 driving trips without any type of pre-determined driving instruction on a wide variety of road scenarios.  We propose an unsupervised learning technique that clusters drivers in different groups, and offers a validation method to test the robustness of clustering in a wide range of experimental settings. The minimal amount of data needed to preserve robust driver clustering is also computed. The presented study provides a new methodology for near-real-time classification of   driver   behavior   in uncontrolled environments.
\end{abstract}

\bigskip


\bigskip

\section{Introduction}

Modern cars are equipped with several hundreds of sensors and electronic control units (ECUs) \cite{Hallac2016} that, beyond guaranteeing an optimal functioning of the engine, provide the driver with more safety, control and entertainment. These almost real­-time data provide information on the car, the driver and the surrounding environment and can be used to study, analyze, predict and understand a large variety of problems, such as traffic congestion, vehicle energy consumption and emissions, urban mobility and drivers' habits \cite{Massaro2017}. 

This huge amount of diverse data has been made available by the CAN bus technology, a serial broadcast bus developed by Robert Bosch in 1986 \cite{Kiencke1986} that allows  communication among the electronic control units devices mounted on the car. CAN technology has become \emph{de facto} a standard in car embedded systems 
providing access to data from an order of several thousands signals, recording at a sub-Hertz frequency information about the car and its surroundings. 

With this technology being implemented in modern cars, the amount and variety of collected data increases and all the aforementioned applications can be extended and improved with respect to the state of art of GPS-based technologies. Data availability is not a restrictive aspect anymore as insights from travels can be collected automatically, without the need to modify the car structure or to specifically design an experiment. Moreover,  in the present research we leverage a data stream in the order of few gygabytes per hour, which represents just a significative sub-sample of all the information travelling on the CAN bus: this amount of data  will only increase with the advent of new autonomous driving cars \cite{Moll}.

\subsection{Driving behavior}
The characterization of driving behavior is not only crucial for accident prevention, as most of car accidents are due to human mishandling 
, but it is also important for designing driving models, which are the core of algorithms that might make the future of self-driving cars possible \cite{Wang2014}. Driving behavior characterization is useful also for car insurance companies to quantify accident risk and provide personalized rates 
State-of-art technology implements models mostly based on GPS location, traveled distance and coarse grained speed profile \cite{Grengs2008, Paefgen2011}. A richer information like the one coming from CAN bus could better characterize human driving behavior and, consequently, accident risk. 

In order to be able to use CAN data to characterize drivers in real application scenarios we need to solve two very challenging problems: (1) providing a methodology for consistently identifying driving behavior in a completely uncontrolled environment, and with very limited knowledge of the surrounding conditions; and (2) minimizing the communication and computational load needed to solve (1). This paper introduces and discusses ideas to tackle these challenges and bring CAN bus based driver characterization closer to reality.


More specifically, the goal of the present research is to extract  features from CAN bus signals and assess to what extent they are useful for finding similarities among drivers using a clustering algorithm. Given the enormous amount of data generated by the CAN bus -- in the order of a few gigabytes of data per hour -- it is not feasible to communicate and process the raw output of the CAN bus in real time to characterize drivers. As such, feasibility of the devised driver characterization methodology is bounded to the definition of a strategy to substantially reduce the amount of data to be processed to perform the driver identification task. Thus, in the second part of the paper we explore different data subsampling methods that allow minimizing data communication between vehicle and infrastructure while guaranteeing robust driver behavior characterization. 

The  paper is organized as follows. Section \ref{sec:data_collection} describes the details of the data collection process and the signals considered. Section \ref{sec:cluster} is devoted to the clustering of the drivers. Section \ref{sec:subsampling} addresses the sampling method question. Finally, section \ref{sec:conclusions} concludes the paper providing a summary of the future research directions.

\subsection{Related work}\label{sec:literature}

In general, research on driving behavior in scientific  literature can be classified according  two  perspectives: (1) the purpose of the research, e.g. driver recognition, maneuver recognition, aggressive or eco-friendly driving detection, \emph{etc}. or (2) the data used for the analyses, i.e. GPS locations, CAN bus data, audio-video data, cellular phone data, car simulator data. 

Early studies have been made with the aim of characterizing driving behavior by building a dynamic model to eventually implement a control system that would react like a human, to be used for example in  self-driving cars.  Models have been proposed to anticipate  the driver actions by few seconds  \cite{Pentland1999} or to predict the  driver's intended cruising speed up to 20 seconds in advance of reaching that speed  \cite{McNew2012}. All these works have been validated using data coming from car simulators. Data acquaired by a simulator have also been used to quantify the drivers' skills \cite{Zhang2010}. 

Some other works, on the other hand, have been conceived to recognize driving maneuvers (e.g. passing, changing lines, turning, starting and stopping) leveraging CAN data: for example, in \cite{Oliver2000} the drivers were asked by an instructor in the vehicle to perform given maneuvers.

Carmona \etal \cite{Carmona2015}, through a novel  hardware tool designed to integrate data from CAN bus, GPS and and an Inertial Measurement Unit (IMU), attempt to classify real-time normal and aggressive driver behavior. The classification was performed in an experiment where 10 drivers have been asked to drive the same route twice, in a normal and aggressive way respectively.

CAN sensors have also been coupled with external devices, designed and mounted specifically on the vehicle for the purpose of the experiment, like  3D cameras for eye monitoring or wereable devices used to collect biomedical signals. These experiments are more \virgolette human-centric'' and are aimed at understanding how drivers' bad habits or distractions are reflected in their way of driving: Choi \etal\cite{Choi2007} and, lately, Li \etal \cite{Li2013} detected and classified distraction tasks  (e.g tuning the radio, interacting with an automatic voice portal) using audio and video data coupled with CAN bus data.  

On the other hand, some works focus on the driver recognition problem, which attempts to distinguish different drivers only by looking at the CAN bus data. Wakita \etal \cite{Wakita}, using data coming from a car simulator, made a comparison between parametric and nonparametric models, concluding that nonparametric approaches perform better in terms of percentages of drivers correctly recognized. Hallac \etal \cite{Hallac2016} leveraged the same database used in this work acheiving a prediction accuracy of 76.9\% for two-driver
classification, and 50.1\% for five drivers. Miyajima \etal \cite{Miyajima2007, Miyajima2006} performed driving recognition modelling on pedal operation patterns acquired by CAN bus sensors by means of a cepstral method, both on a car simulator and on real cars involving 276 drivers. However, the exact setting of the experiment, the type of road the drivers used, and how they have been instructed to drive is not specifically mentioned in the paper. Moreover, the vehicle used for  data collection  (a minivan, \cite{Kawaguchi2004}), equipped with cameras, microphone, computer rack, power suppliers and amplifiers, suggests that the experimental conditions were far from an everyday context in personal driving. 

More recent work uses data coming from mobile phones sensors (accelerometer, gyroscope, magnetometer, GPS, video): in \cite{VanLy2013}, cell phone sensors data have been coupled with CAN bus data as a \virgolette ground truth'' for isolating acceleration, braking and turning events: the problem of driver recognition was addressed, but the experiment  involved only two drivers and reached only 60\% of accuracy. Moreover, mobile phones sensors have been used to detect aggressive \cite{Johnson2011} or drunk \cite{Dai2010} drivers. 

In contrast to the present research, in which normal cars have been used, most of the previously cited works used cars developed in specific projects, like the UTDrive project\footnote{http://www.utdallas.edu/research/utdrive/}  \cite{Angkititrakul2009} or a specificly designed \virgolette vehicle corpora for research'' \cite{Kawaguchi2004, Takeda2011}. Finally, uncontrolled experimental settings have been used in the SHRP2 Naturalistic Driving.\footnote{https://insight.shrp2nds.us/}   

 Study, where driving bahavior has been analized using traditional tecniques (thus not through CAN data) and in another  large experiment called \virgolette EuroFOT'' (European large scale Field Operational Test on in-vehicle system) \footnote{http://www.eurofot-ip.eu/}, where CAN bus data have been used with the only aim of evaluating the impact of 8 different driving assistance systems. 

Comprehensive analyses of driving behavior models, tools and experiments can be found in \cite{Wang2014, Li2013,  Meiring2015}. Summarizing, none of the existing work analyzed usage of CAN bus data for driver classification in a completely uncontrolled and open driving environment. Furthermore, the issue of how to reduce the communication and computational load related to driver classification has, to our best knowledge, never been addressed so far. 

\subsection{Motivations}

As it turns out from the previous section, the main novelty of this paper in the field of human driving behavior analysis is the combination of (1) large number of drivers, (2) completely uncontrolled experimental settings and (3) quantity of data recorded.

 This sets new limits and possibilities to the present research: limits in terms of the variety of the signals acquired, carrying useful information not supported by \virgolette ground truth'',  i.e. information we can consider as \virgolette true'' to which compare the experimental data (for example the \virgolette aggressiveness'' of the driver, his driving skills or his number of incidents). On the other hand, the framework of the present research opens the way to new CAN-based technologies that could find application in real-life scenarios.

\section{Data collection}\label{sec:data_collection}

\subsection{Experimental settings}

The dataset used in the present research has been collected during an experiment carried out by AUDI AG and Audi Electronics Venture. The data collection experiment took place in the city of Ingolstadt (Germany) and involved 64 different drivers, who have not been instructed in any way on the route they had to drive, on the speed or on the behavior they had to follow during the driving. This gives to the present study its unique characteristic of an experiment under uncontrolled testing conditions. A test fleet of ten Audi A3 vehicles was retrofitted with data loggers. This prototype system enables data acquisition for research purposes.

The data collection phase took place in 2014 with a total of 55 days of experiment. Cars were picked up by the drivers in a central deposit and had to be returned within the same day. Each time a user switched on the car engine, the computer registered a new \emph{session}. A total of 1987 sessions have been recorded, and more than 2135 hours of driving data for each of the 2418 sensors have been acquired. Each user drove an average of 31 sessions, whose average duration was 64 minutes. 

CAN bus signals have been recorded on a data logger\footnote{No personal information on the drivers have been recorded.} and processed in a later phase. The sampling is not uniform due to the particular characteristics of the CAN bus and the signals. Therefore, high frequency signals are constantly sampled at 20 Hz, while low frequency sensors reports their data only when there is a change in their value (e.g. rain sensors, seatbelt sensors, etc.) but for the sake of simplicity all the signals considered in the analysis have been resampled at 4 Hz through linear interpolation.

\subsection{Signals selection}

Among the 2418 signals transmitted on the CAN bus, in this work we concentrated the analyses on eight signals:
\begin{itemize}
\item Brake pedal pressure (BRK)
\item Gas pedal position (GAS)
\item Revolutions per minute (R.P.M.)
\item Speed (SPD)
\item Steering wheel angle (S.W.A.)
\item Steering wheel momentum (S.W.M.)
\item Frontal acceleration (F. ACC.)
\item Lateral acceleration (L. ACC.)
\end{itemize}

\begin{figure}
\centering
\includegraphics[width=\columnwidth]{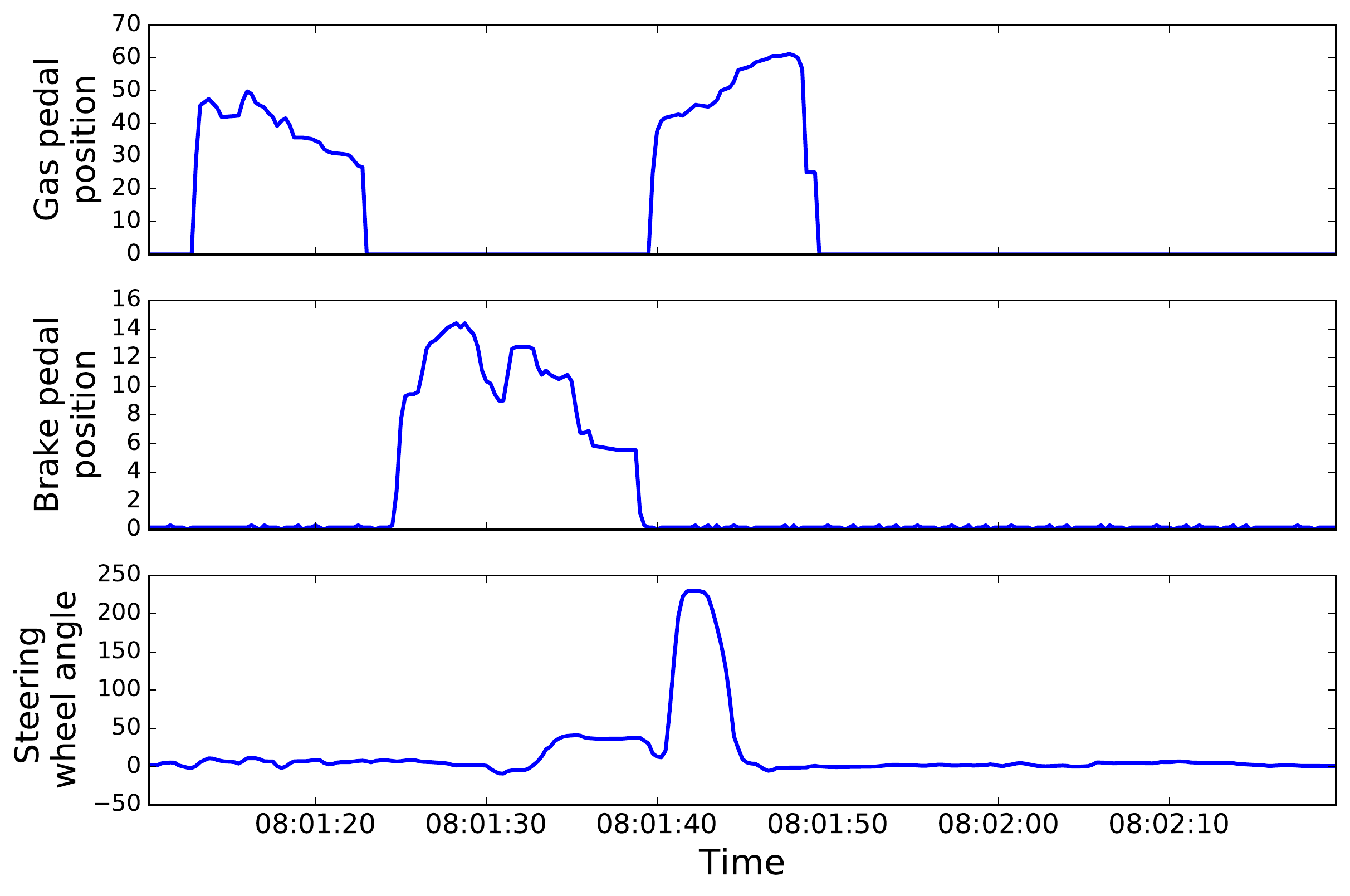} 
\caption{Example of signals acquired by the gas pedal position sensor (top), brake pedal pressure sensor (middle), steering wheel angle sensor (bottom). The three signals have been acquired synchronously.}
\label{fig:signal-samples}
\end{figure}

These signals are directly or, in some cases, indirectly related to the interaction between the driver and the vehicle. For instance, pedals and steering wheel signals directly reflect driver's movements and actions, without any \virgolette transfer function'' between the input (the driver's action) and the output (the signal); some other (speed, rpm and accelerations) represent on a phenomenological point of view quantities that a person can \virgolette feel'' during the driving and could reflect specific driving habits: for example, a driver's attitude to exceed speed limits. An example of the collected signals is reported in Figure \ref{fig:signal-samples}.

\section{Grouping drivers' behavior}\label{sec:cluster}

In this section we propose a methodology that allow us to group in a consistent way the drivers according to common characteristics. 
This methodology is composed of 4 different steps: A) Features extraction, B) Features normalization,  C) Dimensionality reduction and D) Unsupervised Clustering.

\subsection{Feature extraction}\label{sec:features}

Any signal $\textbf{x}$ in the database can be represented as a set of pairs of the type $(x_i, t_i)$, where $i\in\N$ and $t_i$ is the timestamp corresponding to the acquisition of the signal value $x_i$ where $x_i$ is a floating point number. From each considered signals we extract the following $7$ indicators:
\begin{enumerate}
    \item  values of the signal for each sample: $x_i$.
    \item difference quotient (discrete first derivative) of the signal between two consecutive samples: $\frac{x_{i+1}-x_{i}}{t_{i+1}-t_{i}}$. This measure quantifies the intensity of signal variation over time. Let us now define $J$ as the set of indexes for which the values $x_i$ are singular points (local maxima or minima), i.e. $J=\{i:(x_i-x_{i-1})(x_{i+1}-x_i)<0\}$, and by $J_\text{max}\subset J$ the set of only local maxima. Moreover, let us define on those sets a relation $\prec$, where $j\prec k$ means that $j$ is the largest element of the set that precedes $k$, i.e. $j=\max\{i\in J:i<k\}$.
    \item time interval between two singular points: $t_j-t_k, \quad j,k\in J,\ j\prec k$. This feature represents the frequency of its peak points, or in other words the rapidity of variation of the signal when it reaches extreme values.
    \item value of the local maxima: $x_j, j\in J_\text{max}$. This feature provides the intensity of the extreme values of the signal. In a temporal window of one minute and remembering the 4 Hz sampling we define the set of indexes $I_i = \{i-120, \ldots, i+120\}$ and the following.
    \item moving mean, averaging the values $x_i$ over a temporal window of 1 minute: $\frac{1}{240}\sum_{j\in I_i}x_j$.
    \item moving median, the median value of the set $\bigcup_{j\in I_i}x_j$.
    \item moving standard deviation,  the variance of the values  in the set $\bigcup_{j\in I_i}x_j$.
\end{enumerate}

\begin{figure}
\centering
\includegraphics[width=\columnwidth]{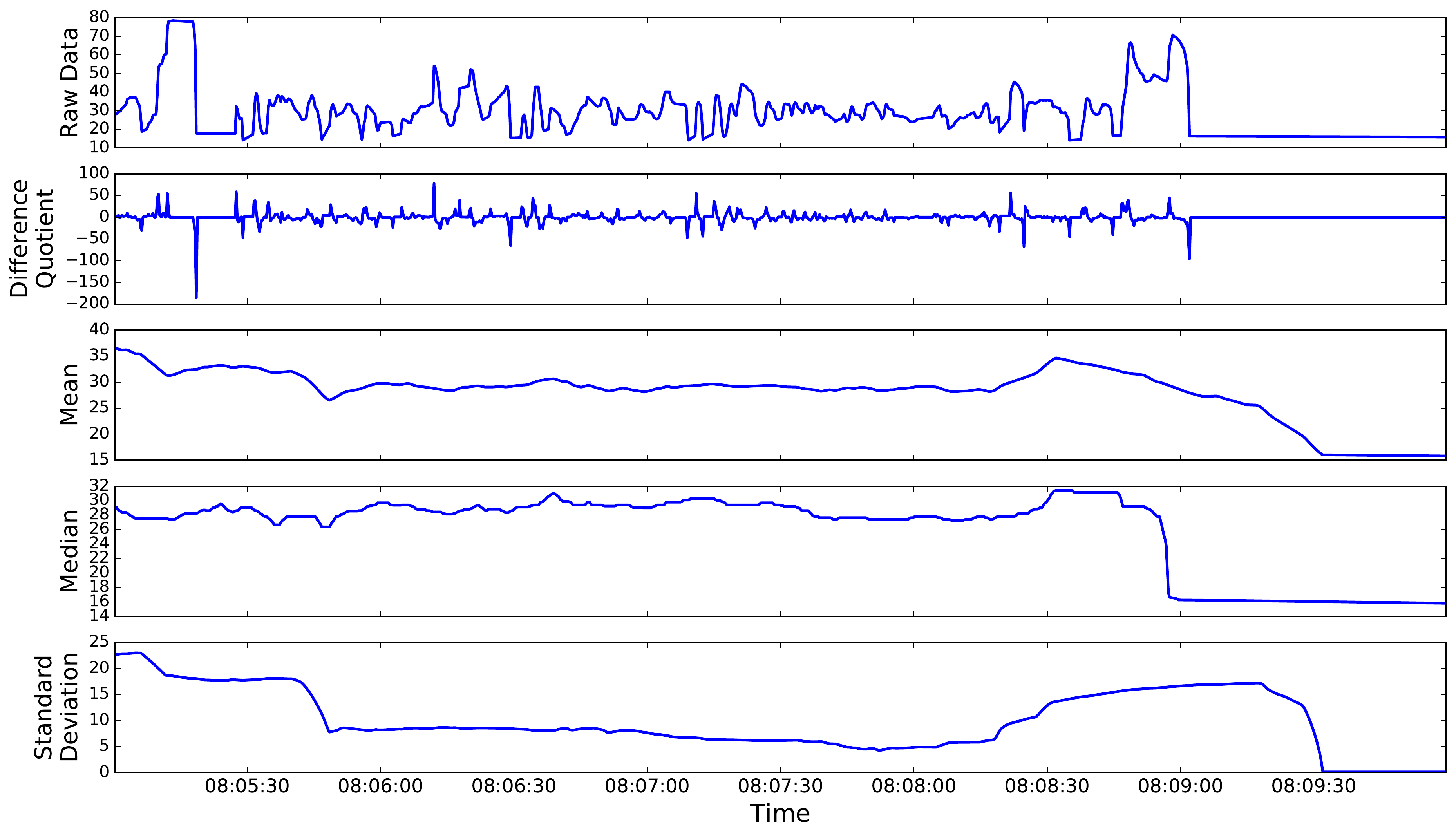} 
\caption{A sample of some of the features extracted from the eight considered signals. In particular, the figure shows the gas pedal angle signal and its difference quotient, mean, median, and standard deviation.}
\label{fig:feature-samples}
\end{figure}

Table \ref{table:Indicators} summarizes the features defined above for a quick reference, while Figure \ref{fig:feature-samples} shows a plot of a sample signal and some of the features.

{\renewcommand{\arraystretch}{1.5}
\begin{table}
\centering 
\begin{small}
\begin{tabular}{cl} 
\hline
\emph{Feature} & \emph{Description} \\ 
\hline
1   & Values of the signal for each sample \\
2   & Difference quotient (discrete first derivative) \\
3   & Time interval between two singular points     \\
4   & Values of the local maxima    \\
5   & Moving mean  \\
6   & Moving median  \\ 
7   & Moving standard deviation   \\ 

\hline
\end{tabular}
\end{small}
\caption{Features definition.} 
\label{table:Indicators} 
\end{table}
}

\subsection{Features normalization}\label{sec:featuresN}

For any given signal $\textbf x$ of floating point type, we denote by $\textbf{w}^{k, u}$  the vector of the feature $k$ for user $u$, obtained by calculating the functions defined above on the vector $\textbf x$,  joining all the sessions  of the same user.  We then normalize each vector  $\textbf{w}^{k, u}$  in the following way.  Outliers removals has done by keeping only the values between the 2\textsuperscript{nd} and 98\textsuperscript{th} percentile. We consider the vector $\textbf{w}^{k, u}$ as a set of statistical samples that are used to build frequency histograms. 

In order to get for each user histograms with the same bins, we define the set
$$W^k = \bigcup_{u\in\mathcal{U} }\bigcup_i\{w^{k,u}_i\}~,$$
where $\mathcal{U}$ is the set of users, 
and partition the interval $[\min W^k, \max W^k]$ into 10 equal intervals\footnote{The number 10 has been chosen after some preliminary analyses. The rationale for choosing the number of bins was to have a sufficient number of bins to well represent the shape of the probability density distribution, but small enough to keep the computation of the machine learning algorithms feasible.} (bins) $b^k_1, \ldots, b^k_{10}$. Then, for each user and for each indicator, the histogram $H^{k,u}$ for the vector $\textbf{w}^{k, u}$ with bins $b^k_1, \ldots, b^k_{10}$ can now be computed, i.e. each bar of the histogram has a value $h^{k,u}_i$ which is the number of items of the vector $\textbf{w}^{k, u}$ belonging to interval $b^k_i$. Finally, all the histograms are normalized, obtaining new values $\tilde h_1, \ldots, \tilde h_{10}$ according to the formula 
$$\tilde h^{k,u}_i = \frac{h^{k,u}_i}{\sum_{j=1}^{10}h^{k,u}_j},$$
so that $\sum_{i=1}^{10}\tilde h^{k,u}_i = 1$.

 According to our definition, features in form of histograms can be interpreted as a discrete version of the sample distributions of the indicator vectors. This definition, along with its probabilistic interpretation, has two main advantages: it allows to perform analyses on objects which have a probabilistic meaning, while on the other  hand it keeps machine learning algorithms relatively simple due to the low dimensionality of the data.

In the following analyses, for data homogeneity we consider users who drove in total at least 10 hours, reducing the number of considered users to 54 from the initial 64.

\subsection{Dimensionality Reduction}\label{par:dataviz}

In this section we use the $K$-means clustering algorithm \cite{Hastie2009} to leverage the features defined in the previous section with the aim of grouping drivers upon common similarities. This is a novel approach in this field and therefore it requires an assessment of the validity of the method in terms of robustness and scalability.

It is worth remarking that the vectors $H^{k,u}$ are 10-dimensional data-points, being them histograms with 10 bins. In order to plot them on bi-dimensional space, therefore, a dimensionality reduction tecnique has to be performed. In this work we use \emph{Principal Component Analysis} (PCA), a well known statistical procedure that decreases the dimensionality of a space projecting it into another one whose dimentions (principal components) are orthogonal to each other and such that the variance of the projected data-points on the principal components is maximized \cite{Hastie2009}. 

\begin{figure}
\centering
\includegraphics[width=1\columnwidth]{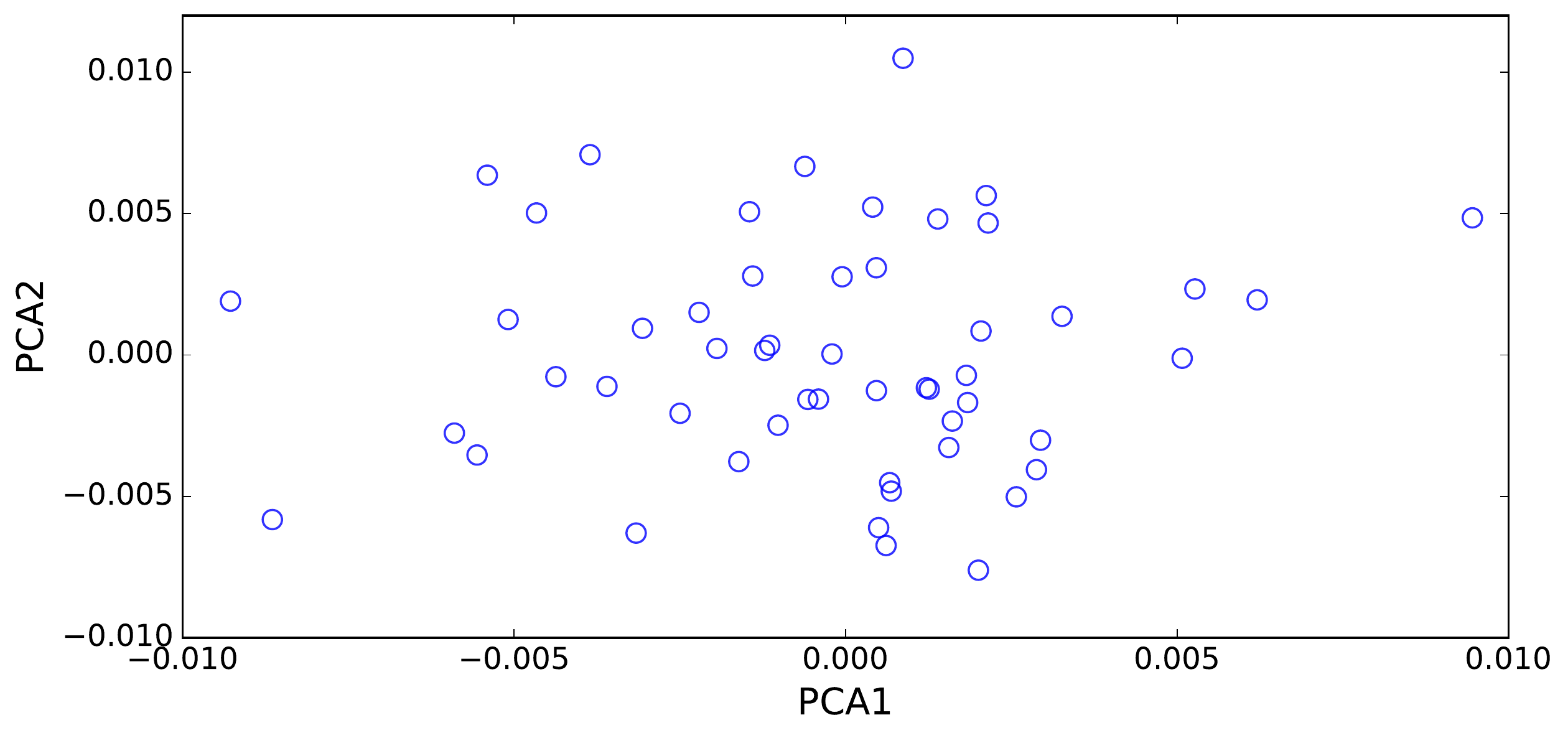}  
\caption{PCA representation for Feature 1 of the gas pedal position signal, where each point represents a different driver.}
\label{fig:PCA}
\end{figure}

Table \ref{tab:eigenvectors} shows that for most of the combinations of signals and features, the first two principal components explain more than 80\% of the total variance of the original high dimensional data. Figure \ref{fig:PCA}, consequently,  reports an example of a bidimensional representations of the features (Feature 1 for the gas pedal signal) where each dot corresponds to a driver. It can be noticed that there are no well separated clusters: this can be expected thinking that human behavior typically varies in a range that forms a continuum. For this reason, the word \virgolette segmentation'' more accurately describes this process than \virgolette clustering'': some common behavior can be identified, while some ``outliers''  slightly deviate from the average.

\begin{table}[]
\centering
\begin{small}
\begin{tabular}{lccccccc}\hline
&\multicolumn{7}{c}{\textit{Features}}\\
                             & 1 & 2 & 3 & 4 & 5 & 6 & 7 \\ \hline
\multicolumn{1}{l}{BRK} & 1.00 & 0.99 & 1.00 & 0.96 & 1.00 & 0.66 & 0.89 \\ 
\multicolumn{1}{l}{GAS} & 0.90 & 0.98 & 0.93 & 0.85 & 0.79 & 0.96 & 0.78 \\ 
\multicolumn{1}{l}{R.P.M.} & 0.61 & 0.95 & 0.57 & 0.78 & 0.70 & 0.98 & 0.73 \\ 
\multicolumn{1}{l}{SPD} & 0.61 & 0.88 & 0.54 & 0.77 & 0.55 & 0.91 & 0.65 \\
\multicolumn{1}{l}{S.W.A.} & 0.92 & 0.99 & 0.92 & 0.97 & 0.95 & 0.97 & 0.80 \\
\multicolumn{1}{l}{S.W.M.} & 0.79 & 0.96 & 0.79 & 0.94 & 0.89 & 0.98 & 0.88 \\
\multicolumn{1}{l}{F. ACC.} & 0.82 & 0.94 & 0.76 & 0.81 & 0.87 & 0.97 & 0.75 \\ 
\multicolumn{1}{l}{L. ACC.} & 0.99 & 0.99 & 0.99 & 1.00 & 1.00 & 0.98 & 0.99 \\ \hline
\end{tabular}
\end{small}
\caption{Total variance of the original data explained by the first two principal components, for each combination of signal and feature.}
\label{tab:eigenvectors}
\end{table}

\subsection{Unsupervised Clustering}\label{par:clusters-comp}

Having no previous information about the drivers and their behavior, it is not known \emph{a priori} the number of different attitudes to be detected and whether a driver is correctly  classified (as opposed, for example, to \cite{Carmona2015}). For instance, we cannot tell which of the datapoints represent \virgolette aggressive", \virgolette dynamic'' or \virgolette eco-friendly'' drivers, as this information is not accessible to us. This remarks motivate the choice of clustering techniques, being part of the \emph{unsupervised learning} approaches to data analysis, used when no previous knowledge on the data is available. In fact, unlike \emph{supervised learning}, the former is an exploratory analysis that does not rely on a \emph{ground truth}, a  concept  identifying the \emph{a priori} known information of the data or the information provided by direct observation, as opposed to information provided by inference.

However, a problem arises when the optimal number of clusters has to be chosen and when the overall quality of the clustering has to be evaluated. Some common techniques try to address this difficulty, for example the plot of SSE (sum of the squared differences between each observation and its group's mean \cite{Hastie2009}) or the shilouette index (a measure of how similar an object is to its own cluster compared to other clusters \cite{Rousseeuw1987}), but as mentioned above in our case clusters are not well separated and those techniques do not provide useful results.

Inspired by the widely used method of cross-validation used in supervised learning, we propose here a new approach for establishing the optimal number of clusters, based on the concept of \virgolette robustness'' of the clustering to the road sampling. In fact, remembering that the clusters are made up of distributions that come from sampled data, the clusters should be invariant to a subsampling of the original data. In other words, comparing the clusters generated by different subsampling of the original data, those clusters should be similar.

\begin{algorithm}[t]
 \caption{$K$-means clustering cross-validation algorithm.}
\label{alg:Xval}
\For{\upshape each feature  $k = 1\ldots 7$}{
 \For{\upshape number of clusters $K = 2\ldots 10$}{
  \For{\upshape number of trials $i = 1\ldots 40$}{
  	\For{\upshape each user $u \in \mathcal{U}$}{
  		randomly permute the elements of vector $\textbf{w}^{k, u}$\;
  		$\textbf{w}_T ^{k, u} =$ first 70\% elements of $\textbf{w}^{k, u}$\;
  		$\textbf{w}_V ^{k, u} =$ last 30\% elements of $\textbf{w}^{k, u}$\;
  		}
compute histograms $\{H_T^{k,u}\}_{u\in\mathcal{U}}$ and $\{H_V^{k,u}\}_{u\in\mathcal{U}}$ as in §\ref{sec:features}\;
$T=\{H_T^{k,u}\}_{u\in\mathcal{U}}$  (training set)\;
  	$V=\{H_V^{k,u}\}_{u\in\mathcal{U}}$  (validation set)\;
   	 $\mathcal C_T = K$-means clustering on $T$\;
	$\mathcal C_V  = K$-means clustering on $V$\;
   $v_i = \text{V-measure}(\mathcal C_T, \mathcal C_V)$\;
   }
   $M_{k,K} =  \text{mean}(\textbf v)$\;
   $S_{k,K} = \text{standard-deviation}(\textbf v)$\;
 }
 }
\end{algorithm}

\begin{figure*}[t]
\centering
\includegraphics[width=2\columnwidth]{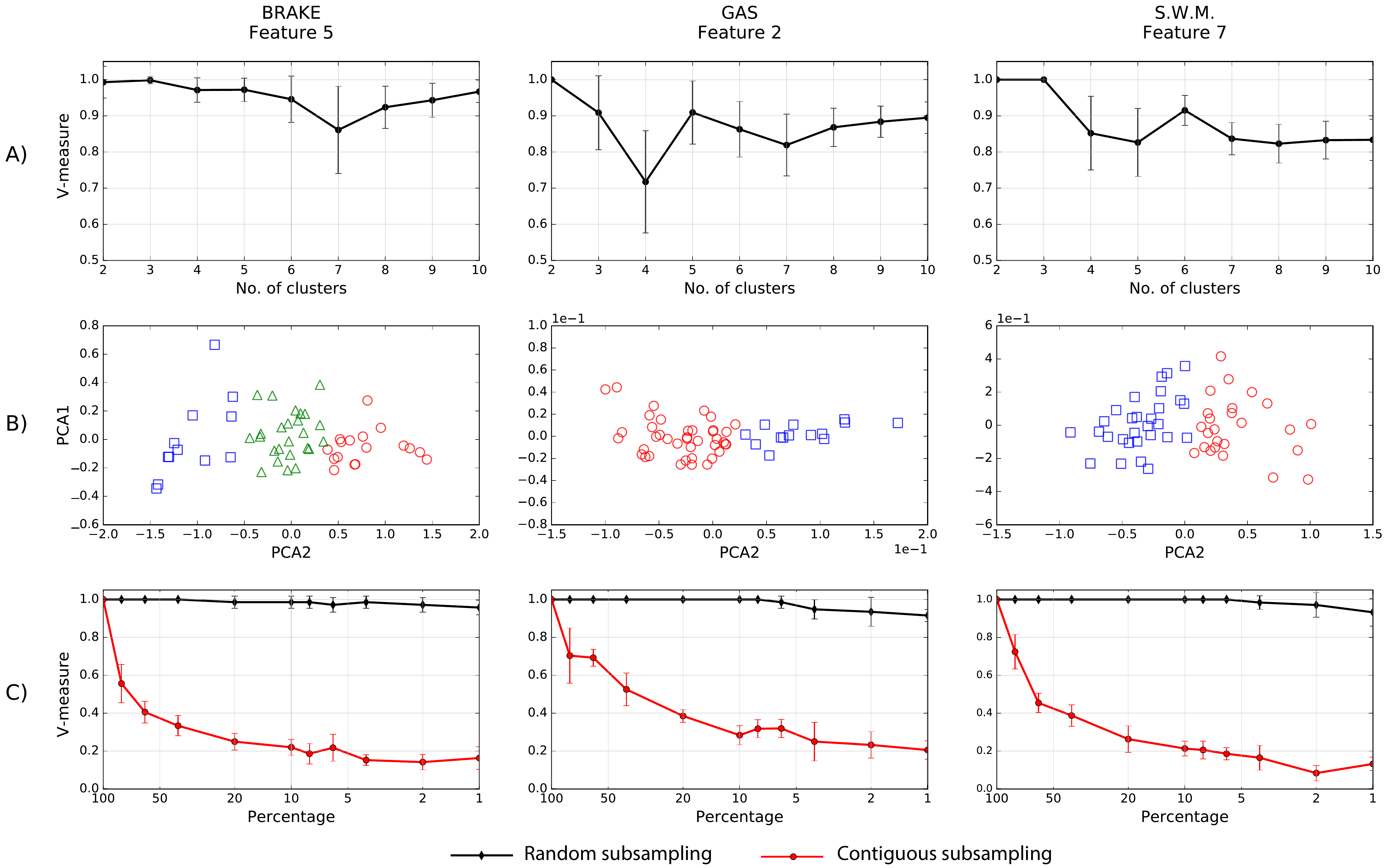} 
\caption{Plot of analyses for selected combinations of signals and features:
(A)  Output of Algorithm \ref{alg:Xval}, plotting the V-measure for different values of $K$;
(B) Drivers clusterings for different signals and features. The $K$-means algorithm has been run on all data in the database and for the optimal values of $K$ as in Table \ref{tab:K-opt}; (C) Subsampling methods: the graphs show the V-measures of the comparisons of the $K$-means clusters generated using all the data in the database, with the clusters generated by a subset of the data (validation set), for different sizes of the validation set (100\%, 50\%, 20\% 10\%, 5\%, 2\%, 1\% of the original data). The clusterings use the optimal values of $K$ as in Table \ref{tab:K-opt}.}
\label{fig:X-val}
\end{figure*}

The method proposed is described in Algorithm \ref{alg:Xval} and can be synthesized as follows. For each user $u$ and for each feature $k$, the vector $\textbf{w}^{k, u}$ is divided into two different vectors: 70\% of its components, taken randomly, form the vector $\textbf{w}_T ^{k, u}$ (\emph{training vectors}), while the other 30\% form the vector $\textbf{w}_V ^{k, u}$ (\emph{validation vectors}). After having computed the histograms for the two sets of vectors, a $K$-means cluster algorithm is performed separately on both the training set and the validation set, producing two different clusterings of the same set of drivers. These two clusterings are then compared using a metric called  \virgolette V-measure'' \cite{Rosenberg2007}, a score ranging from 0 to 1 and evaluating the similarity of the clusterings: if the clusterings are exactly the same (except for permutations on the labels of each cluster) the score is 1, while the score is closer to 0 as the clusterings are more dissimilar. This operations are repeated for a number of clusters $K$ ranging from 2 to 10. Moreover, being the subsampling random, for each value of $K$ the algorithm is repeated 40 times: averages and standard deviations of the scores for each value of $K$ are calculated and lead to plots like the ones in Figure \ref{fig:X-val}A.

The optimal number of $K$ that provides a \virgolette robust'' clusterization is thus defined as the value of $K$ that maximizes the corresponding V-measure in Algorithm \ref{alg:Xval}. Table \ref{tab:K-opt} provides, for each combination of feature and signal, the optimal values together with mean and variance of their corresponding V-measures. In case of ties of the V-measure, the lowest value of $K$ has been considered as the optimal one.

\begin{table*}[t]
\centering
\begin{small}
\begin{tabular}{lccccccc}\hline
&\multicolumn{7}{c}{\textit{Features}}\\
                             & 1 & 2 & 3 & 4 & 5 & 6 & 7 \\ \hline
\multicolumn{1}{l}{BRAKE} & 2 (0.95, 0.11) & 4 (0.99, 0.01) & {2 (1.00, 0.00)} & 5 (1.00, 0.01) & 3 (1.00, 0.01) & 3 (0.95, 0.05) & 2 (0.92, 0.07) \\
\multicolumn{1}{l}{GAS} & 2 (0.96, 0.06) & {2 (1.00, 0.00)} & 2 (0.93, 0.06) & 4 (0.98, 0.03) & {2 (1.00, 0.00)} & 2 (0.99, 0.03) & 2 (0.99, 0.03) \\
\multicolumn{1}{l}{R.P.M.} & 3 (0.99, 0.02) & 2 (0.98, 0.05) & 2 (0.85, 0.06) & {2 (1.00, 0.00)} & {2 (1.00, 0.00)} & 6 (0.71, 0.06) & 2 (0.92, 0.08) \\
\multicolumn{1}{l}{SPEED} & {2 (1.00, 0.00)} & 2 (1.00, 0.02) & 3 (0.81, 0.12) & 2 (0.98, 0.05) & 2 (0.93, 0.06) & 6 (0.72, 0.04) & 2 (0.86, 0.09) \\
\multicolumn{1}{l}{S.W.A.} & 2 (0.98, 0.05) & 5 (0.99, 0.02) & 4 (0.78, 0.08) & 2 (0.99, 0.09) & {4 (1.00, 0.00)} & 2 (0.92, 0.14) & 3 (0.97, 0.05) \\
\multicolumn{1}{l}{S.W.M.} & {3 (1.00, 0.00)} & 2 (0.96, 0.06) & 4 (0.91, 0.05) & 2 (1.00, 0.02) & 2 (0.92, 0.09) & 2 (0.96, 0.06) & {2 (1.00, 0.00)} \\
\multicolumn{1}{l}{F.ACC.} & 4 (0.98, 0.05) & 6 (0.93, 0.06) & 2 (0.88, 0.09) & 5 (0.87, 0.07) & 2 (0.98, 0.05) & 2 (0.82, 0.09) & {2 (1.00, 0.00)} \\
\multicolumn{1}{l}{L.ACC.} & 3 {(0.99, 0.04)} & 2 (0.83, 0.09) & 2 (0.86, 0.10) & 2 (0.92, 0.12) & 2 (0.94, 0.08) & 2 (0.80, 0.10) & 2 (0.97, 0.08) \\ \hline
\end{tabular}
\end{small}
\caption{Optimal number of clusters for each combination of feature and signal as a result of the cross-validation process described in section \ref{par:clusters-comp}. In brackets, the value of mean and standard deviation referred to the optimal value as in Algorithm \ref{alg:Xval} .}
\label{tab:K-opt} 
\end{table*}

Results clearly show that there are some numbers of clusters that separate users in a better way in terms of \virgolette robustness''. For example, feature 2 for the gas pedal position separates drivers in two different groups, which keep exactly the same in all the 40 repetitions of the cross-validation algorithm, whilst it is not the same for $K=4$.

Overall, some features and some signals perform better than other: the brake pressure signal is the one with most promising results, followed by the gas pedal position and the steering wheel. This is a first important result, as it confirms what has been already found in the literature with data from an unstructured experiment \cite{Miyajima2007}.

Finally, Figure \ref{fig:X-val}B reports the results of the $K$-means clustering for a selection of signals (see Figure \ref{fig:clusters_all} in the Appendix for a comprehensive chart), with values of $K$ as in Table \ref{tab:K-opt}.

\section{Dataset reduction}\label{sec:subsampling}

Once we have verified that a consistent, robust clustering of drivers is possible also in completely uncontrolled, open traffic conditions, we tackle the second fundamental aspect for real-life application: the best sampling method and the minimum amount of data required to provide consistent results. In fact, state-of-art technology in car communication uses mobile connectivity to stream data from the car to the server where they are processed, and given the massive volume of the sampled data it is crucial to investigate a lower-bound for this data communication. We compare two methods that involve different spatiotemporal sampling of the data and we study the quality of the clustering with different quantities of analyzed data.

The subsampling of the vectors $\textbf{w}^{k, u}$ presented in Section \ref{par:clusters-comp} is completely random and does not consider any spatial or temporal dimension: in other words, it is an \emph{independent subsampling}. We compare it with a different subsampling strategy, which we call \emph{contiguous subsampling}, a subsampling conditioned to spatial contiguity defined as follows. Given the vector $\textbf{w}^{k, u}$ of dimension $d$, a random number  $r\in \N$ is extracted uniformly in the interval $[1, d]$. Setting $l=\lfloor pd\rfloor$, where $p\in(0,1)$ is the percentage of the elements to be subsampled, the vector $\textbf{w}^{k, u}_S$ is constructed considering the elements of $\textbf{w}^{k, u}$ with indexes from $r$ to $(r+l)\mod d$. In other words, the vector is subsampled taking, starting from a random element, its $l$ consecutive elements, considering the vector with a circular structure.  

For each of the two subsampling strategies defined, we propose an analysis that compares the clusterizations generated in two different ways: in the first, drivers are clustered upon all the data in the dataset, i.e. data coming from all the roads they have driven on; in the second, drivers are clustered upon only a portion of the data acquired. In this way, the first clustering can be considered somehow as a ground truth (being the result of all the data available to us), while the second is the result of a partial subsampling.


Figure \ref{fig:X-val}C reports the results of the V-measure comparisons of the clusterings generated using all the data in the database with the clusterings generated by a subset of the data, for different sizes of subsets and for the two aforementioned subsampling methods. Every subsampling has been repeated 40 times with different random numbers and the $K$-means clusterings have been performed for each feature with the optimal value of $K$ found earlier. 

Results clearly show that the independent subsampling strategy performs better than the contiguous one, and for some features and signals it is possible to reduce the original dataset by a factor of 100 without impairing clustering performance. A comprehensive chart for all the combinations of signals and features can be found in Figure \ref{fig:howlow} in the Appendix.

\section{Conclusions} \label{sec:conclusions}

In this paper, the problem of driving behavior analysis has been studied from a new point of view, that bridges the gap between driving behavior studies through uncontrolled experiments -- leveraging only the GPS signal -- and studies exploiting CAN bus data through very controlled experiments. This work proposes a methodology for delineating  similarities among drivers using data collected in a completely uncontrolled experiment, through a clustering algorithm performed on seven different features of eight signals recorded by CAN bus sensors, with a distributional approach. Moreover, it has been shown that, by properly choosing the subsampling strategy, it is possible to reduce the size of the dataset of as much as 99\% without impairing clustering performance.

\subsection{Discussion}

Given the almost ontological question of what driver behavior is, this work attempts to define it through a data-driven approach. Without any external knowledge (ground truth), though, it is unclear how to define the boundary between the performance of the proposed method and the fuzziness and the unpredictability of human behavior. However, the promising results obtained in this study suggest that the present approach could be considered as a methodology for testing new signals, features and clustering methods which, coupled with additional field knowledge, may lead to pragmatic interpretations of the different clusters in terms of physical and behavioral characterization of driving styles.

It is important also to outline some limitations of this work: the number of users, 64 later reduced to 53 for data homogeneity reasons, likely does not offer a rich enough variety of driving behaviors to enable a comprehensive identification of common attitudes and outliers. Finally, an aspect that needs further investigation is the interaction of the different indicators and the signals directly in the clustering process.


\subsection{Applications and future work}\label{par:applications}

This paper projects the problem of driving behavior characterization using CAN bus technology from a research-oriented approach into an application-oriented technology that opens the way to wide scale and real-time implementations. In fact, as mentioned, the presence of the CAN bus data in almost every car could scale-up any possible application in a very broad and cost-effective way.

Car insurance companies, for example, are interested in assessing the risk of accidents for each user based on real data coming from their driving sessions. Users segmentation in fact, to the best of our knowledge, today is only performed -- besides the accidents history -- on general information like the geographical location, distance traveled, and velocity. More sophisticated concepts like \virgolette aggressiveness'' or \virgolette nervousness'' could be fully characterized. However, in order to do so, further studies have to be performed, comparing the insurance companies drivers' profiles with the clustering obtained in this work, allowing their characterization based on a ground truth.

Another application is driver recognition, aiming to recognize a driver only upon the CAN bus data. This driver \virgolette fingerprint'', already studied \cite{Enev2016} but never tested in an uncontrolled experimental scenario, could let the car itself to identify the driver for security reasons or adapting settings for comfort or efficiency optimization.

Finally, integration of this modeling technique with physical  detection technologies including sonar devices, stereo cameras, lasers and radar would allow to better understand and model driver behaviors, to improve the development of self driving cars and to have safer road networks.

\bigskip

\footnotesize

\textbf{Privacy disclamier}. The data reported herein was collected during experiments performed with drivers who were hired and were explicitly informed of the data collection process. In case the presented methodology should be used with consumer vehicles, it is fundamental to properly inform the customer about usage of data and the purpose of the collection. This needs to be done in order to comply with data privacy laws and regulations, but also to support customers' awareness and self-determination -- especially in cases where the realization of an application requires providing personal data to third parties. It is the decision of the customer based on a declaration of consent, if personal data may be collected and for which purpose it may be used.

\normalsize

\bibliographystyle{IEEEtran}
\input{AUDI_paper.bbl}

\begin{landscape}
\begin{figure}[tbh]
\centering
\includegraphics[width=0.94\columnwidth]{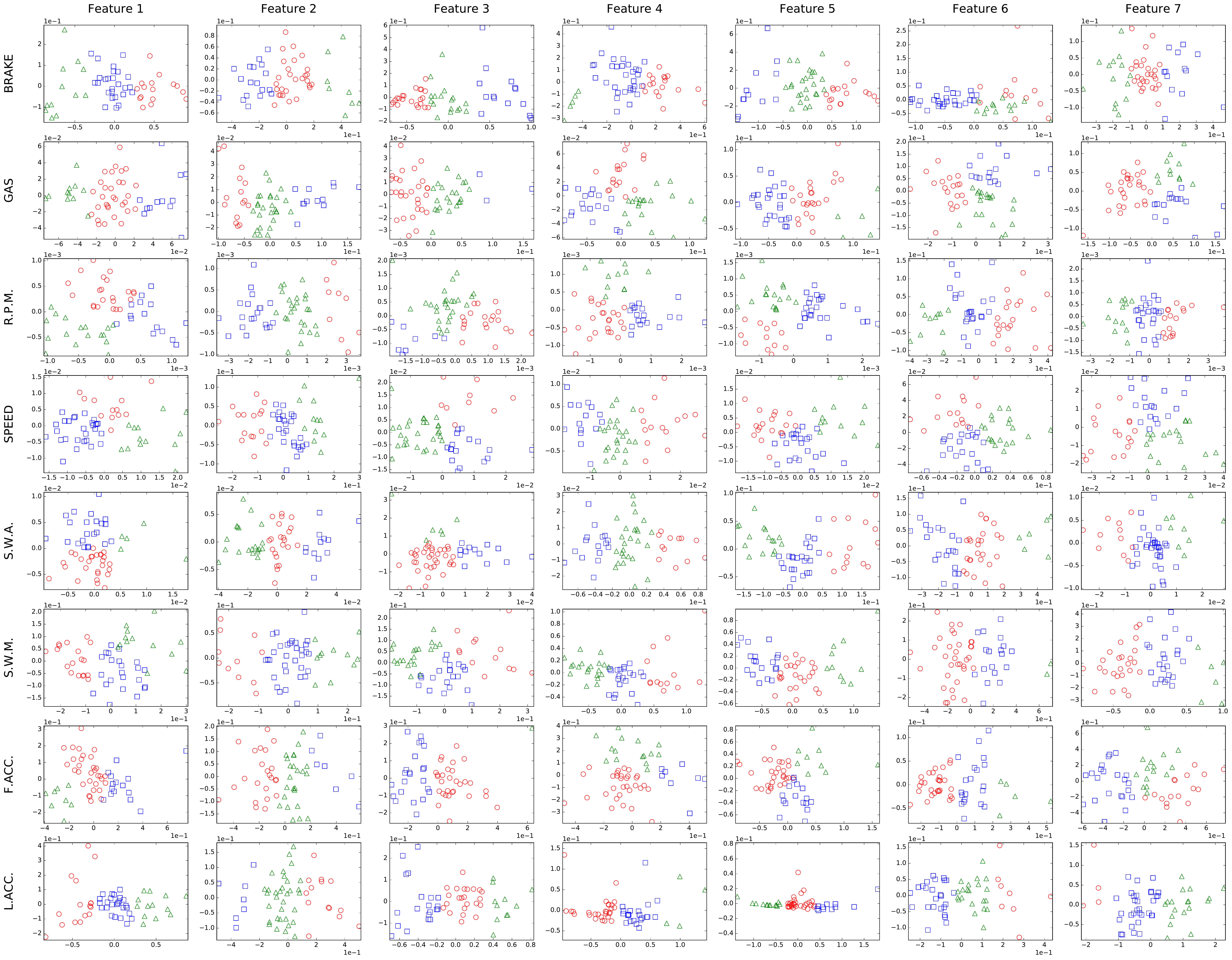}
\caption{Drivers clusterings for different signals and features. The $K$-means algorithm has been run on all data in the database and for the optimal values of $K$ as in Table \ref{tab:K-opt}}
\label{fig:clusters_all}
\end{figure}
\end{landscape}

\begin{landscape}
\begin{figure}[tbh]
\centering
\includegraphics[width=0.94\columnwidth]{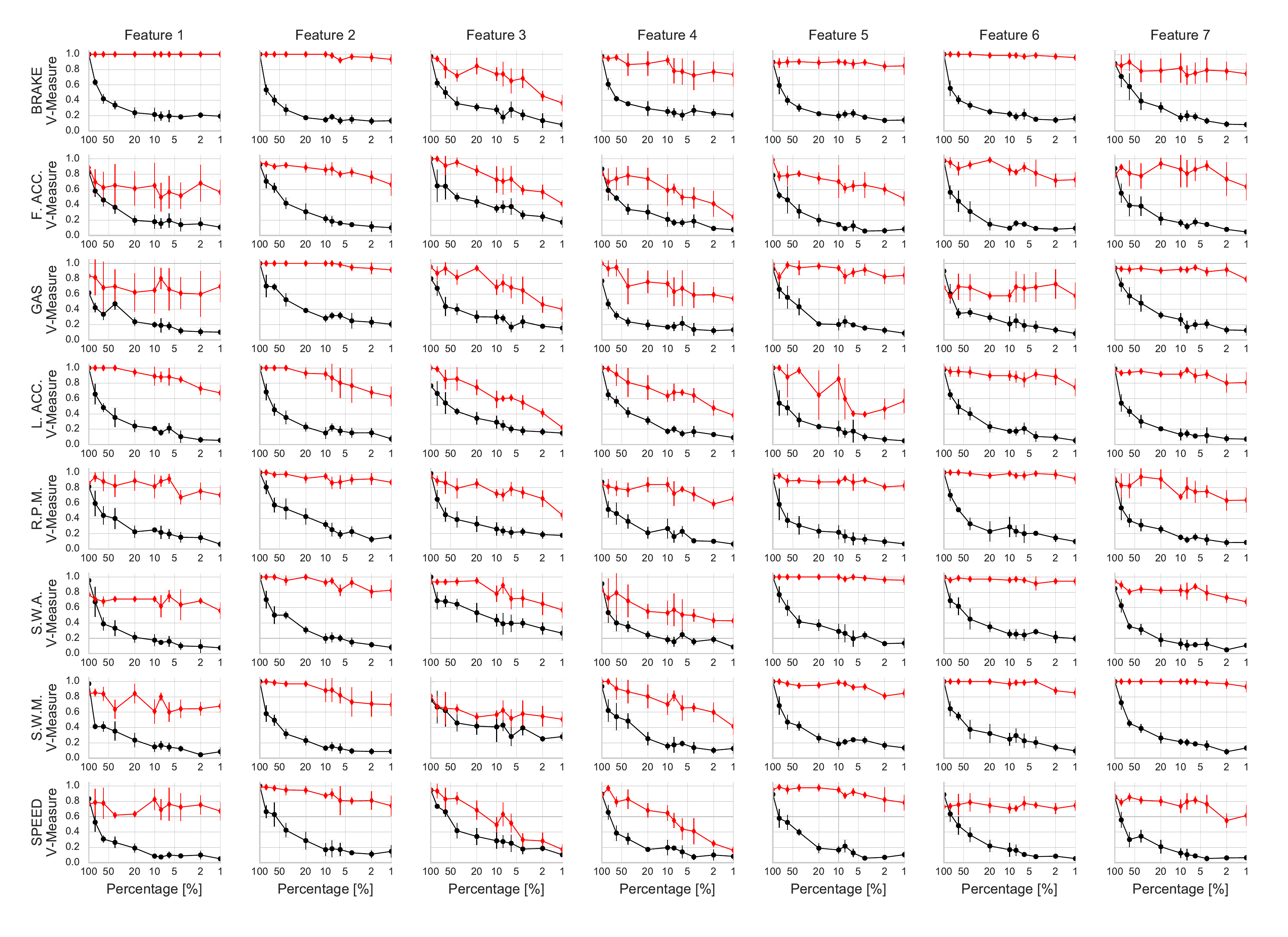}
\caption{Comparison of different subsampling methods: \emph{independent subsampling} (red line, diamonds) and \emph{contiguous subsampling} (black line, circles). V-measures of the comparisons of the $K$-means clusters generated using all the data in the database, with the clusters generated by a subset of the data (validation set), for different sizes of the validation set (100\%, 50\%, 20\% 10\%, 5\%, 2\%, 1\% of the original data). The clusterings use the optimal values of $K$ as in Table \ref{tab:K-opt}.}
\label{fig:howlow}
\end{figure}
\end{landscape}

\footnotesize

\textbf{Umberto Fugiglando} is a Research Fellow at MIT Senseable City Lab. He received his Bachelor degree (2013) and Master degree (2016) in Applied Mathematics from Politecnico di Torino (Itlay), with a thesis on driving behavior. He is also a ASP Alta Scuola Politecnica fellow and he has spent a semester at KTH Royal Institute of Technology in Stockholm (Sweden). His research interests are in the area of digital technology and data science with applications to mobility, acoustics and human behavior characterization. 

\bigskip

\textbf{Paolo Santi} is Research Scientist at MIT Senseable City Lab where he leads the MIT/Fraunhofer Ambient Mobility initiative, and a Senior Research at the Istituto di Informatica e Telematica, CNR, Pisa. Dr. Santi holds a "Laurea" degree and PhD in computer science from the University of Pisa, Italy. Dr. Santi is a member of the IEEE Computer Society and has recently been recognized as Distinguished Scientist by the Association for Computing Machinery. His research interest is in the modeling and analysis of complex systems ranging from wireless multi hop networks to sensor and vehicular networks and, more recently, smart mobility and intelligent transportation systems. In these fields, he has contributed more than 120 scientific papers and two books. Dr. Santi has been involved in the technical and organizing committee of several conferences in the field, and he is/has been an Associate Editor of the IEEE Transactions on Mobile Computing, the IEEE Transactions on Parallel and Distributed Systems, and Computer Networks. Dr. Santi was Guest Editor of the Proceeding of the IEEE special issue on Vehicular Communications: Ubiquitous Networks for Sustainable Mobility in 2011, to which he also contributed a paper.

\bigskip

\textbf{Emanuele Massaro}, PhD is a Postdoctoral Research Fellow at the MIT Senseable City Lab.  He received both his Bachelor (2006) and his Master (2009) in Environmental Engineering from the University of Florence (Italy). He then received his PhD in Complex Systems and Nonlinear Dynamics in 2014 from the Department of Information Engineering and Department of Physics and Astronomy at the University of Florence. He came to the United States in March 2014 to conduct his postdoctoral research where he worked for one year as Postdoctoral Associate at the Department of Civil and Environmental Engineering Carnegie Mellon University and also as a contractor for the Risk and Decision Science Team of US Army Corps of Engineer. He joined the Massachusetts Institute of Technology in March 2015. His broad research interests are in the areas of socio-technical systems and computational social science: he aims to understand the theory of, and quantify the interplay among physical infrastructures, information, and human (societal) activities.

\bigskip

\textbf{Sebastiano Milardo} received his Bachelor degree in 2011 and his Master degree in  2013, both 
in  Computer  Engineering  from  the  University  of  Catania.  From  January  2014  to  April  2015  he worked in the Italian National Consortium of Telecommunications (CNIT), as Researcher within the  NEWCOM\#  and  SIGMA  Projects.  Since  2015  he is currently  a  Ph.D.  student  in  Information  and Communication  Technologies  at  the  University  of  Palermo.  His  research  interests  include Software  Defined  Networking,  Sensor  Networks,  network  protocols  for  the  Internet  of  Things and Big Data analysis.

\bigskip

\textbf{Kacem Abida}, PhD, is a senior engineer at the Volkswagen Group of America Electronics Research Lab (ERL). Dr. Abida is currently leading the big data projects at ERL. He holds a PhD degree in Electrical and Computer Engineering from the University of Waterloo, Canada. His areas of interest include speech and natural language technologies, as well as machine learning based big data analytics.

\bigskip

\textbf{Rainer Stahlmann} received his diploma in electrical engineering and computer science from University of Applied Sciences Ingolstadt, Germany, in 2009.
Since then he has been working for AUDI AG in Ingolstadt, Germany, where he is currently in the Department of Data Strategy and Analytic Services.
In cooperation with the Chair for Computer Networks and Communication Systems at University of Erlangen, Germany, he is working toward his Ph.D. degree.
His research is focused on vehicular data processing and analytics as well as on technical evaluation of V2X communication systems.

\bigskip

Dr. Ing. \textbf{Florian Netter }received his diploma in mechanical engineering from the technical university of Munich, Germany, in 2010. Since then he has been working for AUDI AG in Ingolstadt, Germany, where he received his Ph.D. degree in cooperation with the Karlsruhe Institute of Technology, Germany, in 2015. During his research he focused on complexity adaptation of simulation models in entire system simulations to identify quantification attributes for a high goodness of fit and in spite of increasing computational power still maintaining a short simulation period. Currently he is working at AUDI AG in the Department for Platform Development and Data Analytics taking care of vehicular data stream processing in cloud computing environments.

\bigskip

\textbf{Carlo Ratti} is the founder and Director of the MIT Senseable City Lab. An architect and engineer by training, Carlo Ratti practices in Italy and teaches at the Massachusetts Institute of Technology. He graduated from the Politecnico di Torino and the École Nationale des Ponts et Chaussées in Paris, and later earned his MPhil and PhD at the University of Cambridge, UK. Ratti has co-authored over 200 publications and holds several patents. His work has been exhibited worldwide at venues such as the Venice Biennale, the Design Museum Barcelona, the Science Museum in London, GAFTA in San Francisco and The Museum of Modern Art in New York. His Digital Water Pavilion at the 2008 World Expo was hailed by Time Magazine as one of the Best Inventions of the Year. He has been included in Esquire Magazine Best and Brightest list, in Blueprint Magazine 25 People Who Will Change the World of Design, and in Forbes Magazine Names You Need To Know in 2011. Ratti was a presenter at TED 2011 and is serving as a member of the World Economic Forum Global Agenda Council for Urban Management. He is a regular contributor to the architecture magazine Domus and the Italian newspaper Il Sole 24 Ore. He has also written as an op-ed contributor for BBC, La Stampa, Scientific American and The New York Times.

\end{document}

%% file: newcommands.tex
\newcommand{\N}{\mathbb{N}}

\newcommand{\virgolette}{``}

\renewcommand{\phi}{\varphi}